\documentclass[final,5p,times,twocolumn]{elsarticle}

\usepackage{amssymb}
\usepackage{lipsum}
\usepackage{graphicx}
\usepackage{amsmath}
\usepackage{pgfplots}
\usepackage{caption}
\usepackage{subcaption}
\usepackage{tikz}
\usepackage{comment}
\usepackage{float}
\usepackage{fancyhdr}
\usetikzlibrary{shapes.geometric, arrows}
\tikzstyle{PNode} = [circle, rounded corners, minimum width=1cm, minimum height=1cm, text centered, draw=black, fill=red!30]
\tikzstyle{CNode} = [diamond, rounded corners, minimum width=1cm, minimum height=1cm, text centered, text width = 1cm, draw=black, fill=blue!30]
\tikzstyle{LNode} = [rectangle, rounded corners, minimum width=1cm, minimum height=1cm, text centered, text width = 1cm, draw=black, fill=green!30]
\tikzstyle{arrow} = [thick, ->, >=stealth]

\usepackage{fancyhdr}

\usepackage{xpatch} 
\xpatchcmd{\MaketitleBox}{\hrule}{}{}{}
\xpatchcmd{\MaketitleBox}{\hrule}{}{}{}

\makeatletter
\def\ps@pprintTitle{%
  \let\@oddhead\@empty
  \let\@evenhead\@empty
  \def\@oddfoot{\reset@font\hfil\thepage\hfil}
  \let\@evenfoot\@oddfoot
}
\makeatother

\begin{document}

\begin{frontmatter}

\title{Combustion Condition Identification using a Decision tree based Machine Learning Algorithm Applied to a Model Can Combustor with High Shear Swirl Injector}
\pagestyle{fancy}

\author[first]{PK Archhith}
\address[first]{Department of Mechanical Engineering, Indian Institute of Technology Madras, Chennai-600036, India. }
\author[second]{SK Thirumalaikumaran}
\author[second]{Balasundaram Mohan}

\author[second,third]{Saptharshi Basu$^{*,}$}

\address[second]{Department of Mechanical Engineering, Indian Institute of Science Bangalore, Karnataka-560012, India. }
\address[third]{Interdisciplinary Centre for Energy Research, Indian Institute of Science Bangalore, Karnataka-560012, India. \\$^*$Corresponding author: e-mail: sbasu@iisc.ac.in }

\end{frontmatter}

\fancypagestyle{pprintTitle}{
\renewcommand{\headrulewidth}{0.5pt}
\fancyhead[C]{ \textit{5$^{th}$ National Aerospace propulsion conference at IIT Madras, India, 20-22 Jan 2025}}
}

\pagestyle{fancy}
\fancyhf{}
\fancyhead[C]{\textit{5$^{th}$ National Aerospace propulsion conference at IIT Madras, India, 20-22 Jan 2025}}
\renewcommand{\headrulewidth}{0.5pt}




\hspace{-0.3cm}\textbf{Keywords:} Combustion classification, Gas turbine combustion, Combustion instability, Machine learning, Decision tree. \\

\hspace{-0.3cm}\textbf{Abstract:} Combustion is the primary process in gas turbine engines, where there is a need for efficient air-fuel mixing to enhance performance. High-shear swirl injectors are commonly used to improve fuel atomization and mixing, which are key factors in determining combustion efficiency and emissions. However, under certain conditions, combustors can experience thermoacoustic instability. In this study, a decision tree-based machine learning algorithm is used to classify combustion conditions by analyzing acoustic pressure and high-speed flame imaging from a counter-rotating high-shear swirl injector of a single can combustor fueled by methane. With a constant Reynolds number and varying equivalence ratios, the combustor exhibits both stable and unstable states. Characteristic features are extracted from the data using time series analysis, providing insight into combustion dynamics. The trained supervised machine learning model accurately classifies stable and unstable operations, demonstrating effective prediction of combustion conditions within the studied parameter range.

\section{INTRODUCTION}
\label{introduction}

Stable combustion is essential for fuel efficiency, performance, durability, and low emissions in gas turbine engines. Combustion instability, often seen as thermoacoustic fluctuations, causes cyclic thermal and mechanical stress, potentially leading to engine failure. Predicting these instabilities early is crucial, for this purpose machine learning models were utilized in recent literature and summarized below.

Fichera et al. \cite{ficheraclustering} analyzed the dynamic behavior of lean premixed gas turbine combustors by classifying experimental data using the geometrical properties of phase space attractors, reconstructed from heat release rates. They used unsupervised Kohonen neural networks to cluster operating conditions, successfully categorizing them into stable, unstable, and critically unstable states. Sarkar et al. \cite{sarkarearly} combined Deep Belief Networks (DBN) with Symbolic Time Series Analysis (STSA) for early detection of thermoacoustic instability in gas turbine combustors, showing improved class separability and detection accuracy compared to traditional methods. Another study \cite{akintayoearly} proposed a selective convolutional autoencoder-based deep neural network to predict thermoacoustic instability using high-speed flame images, with the model trained to identify both stable and unstable states. Wang et al. \cite{wangdeep} developed a deep learning model using DNN and CNN frameworks can simultaneously predict combustion states and heat release rates with $99.91\%$ accuracy, demonstrating significant industrial potential. Another study \cite{zhuconvolutional} used CNNs to extract features from acoustic pressure data in a supersonic combustor, outperforming other models in classifying combustion conditions. Hernandez-Rivera et al. \cite{hernandezdetection} used nonlinear time-series analysis of pressure fluctuations to calculate recurrence plot indices, which effectively predicted the transition from combustion noise to instability.

In \cite{gangopadhyaydeep}, the authors used premixed bluff-body stabilized flame images to classify combustion conditions during instability. A CNN was trained to recognize spatial patterns from high-speed flame image sequences, with instability labeled using acoustic pressure data. To capture temporal correlations, they employed Long term short Memory Recurrence Neural Networks (LSTM-RNM), resulting in a binary classification framework combining CNN and LSTM. Wang et al. \cite{wangpattern} developed a pattern recognition model using logarithmic entropy multi-threshold segmentation and fuzzy pattern recognition to classify abnormal combustor conditions, outperforming other methods like self-organizing maps and support vector machines. Han et al. \cite{hancombustion} used a stacked sparse autoencoder-based deep neural network to extract features from unlabeled flame images and combined with loss function to improve training efficiency, achieving superior prediction accuracy. McCartney et al. \cite{mccartneyonline} enhanced thermoacoustic instability prediction with supervised machine learning on dynamic pressure readings, using Hidden Markov Models and Automated Machine Learning to restore predictive power lost by traditional tools like the Hurst exponent and Auto-Regressive models. Zhou et al. \cite{zhoumonitoring} used deep learning to monitor combustion instabilities via time-averaged flame images, designing a CNN called BASIS Image Monitor (BIM) that achieved $99\%$ accuracy in predicting thermoacoustic states and visualized statistical links between flame images and stability using Class Activation Maps.

Jiang et al. \cite{jiangcombustion} developed a decision tree-based method to identify combustion conditions in coal-fired kilns using flame video intensity. By constructing a phase space from the intensity sequence and extracting trajectory evolution and morphology distribution features, the method improved efficiency by over $5\%$ compared to existing approaches. Zhang et al. \cite{zhangexploring} employed a neural network to predict combustion instability, simplifying it with Active Subspaces (AS) to reduce training time while maintaining accuracy. The model uses historical data and system parameters, focusing on the most significant variations captured by AS. In \cite{ahnlongitudinal}, the authors varied the equivalence ratio and observed significant changes in combustor modal dynamics. They used Jensen-Shannon complexity and permutation entropy to classify these dynamics via a k-medoids clustering algorithm, accurately distinguishing between various modes, including stable oscillations and azimuthal modes. Fu et al. \cite{fuearly} applied recurrence quantification and multi-fractal analysis to detect transitions from stable to unstable operation in high-pressure combustors. Using indicators like the Hurst exponent and recurrence rate, their logistic regression and support vector machine models predicted the onset of instability several hundred milliseconds in advance.

It is clear from the literature that various measures and tools were used to predict the combustion condition in the past using acoustic pressure, heat release rate, and high speed flame images in combustor geometries ranging from academic scale to industrial relevance. In the present study, we apply the time series analysis based measures in the literature to high shear counter swirl model gas turbine combustor. This work is a preliminary investigation carried out to identify the combustion condition prior to more practical cases which will be reported in the near future. 
\begin{figure}[h]
    \centering
    \includegraphics[width = 0.5\textwidth]{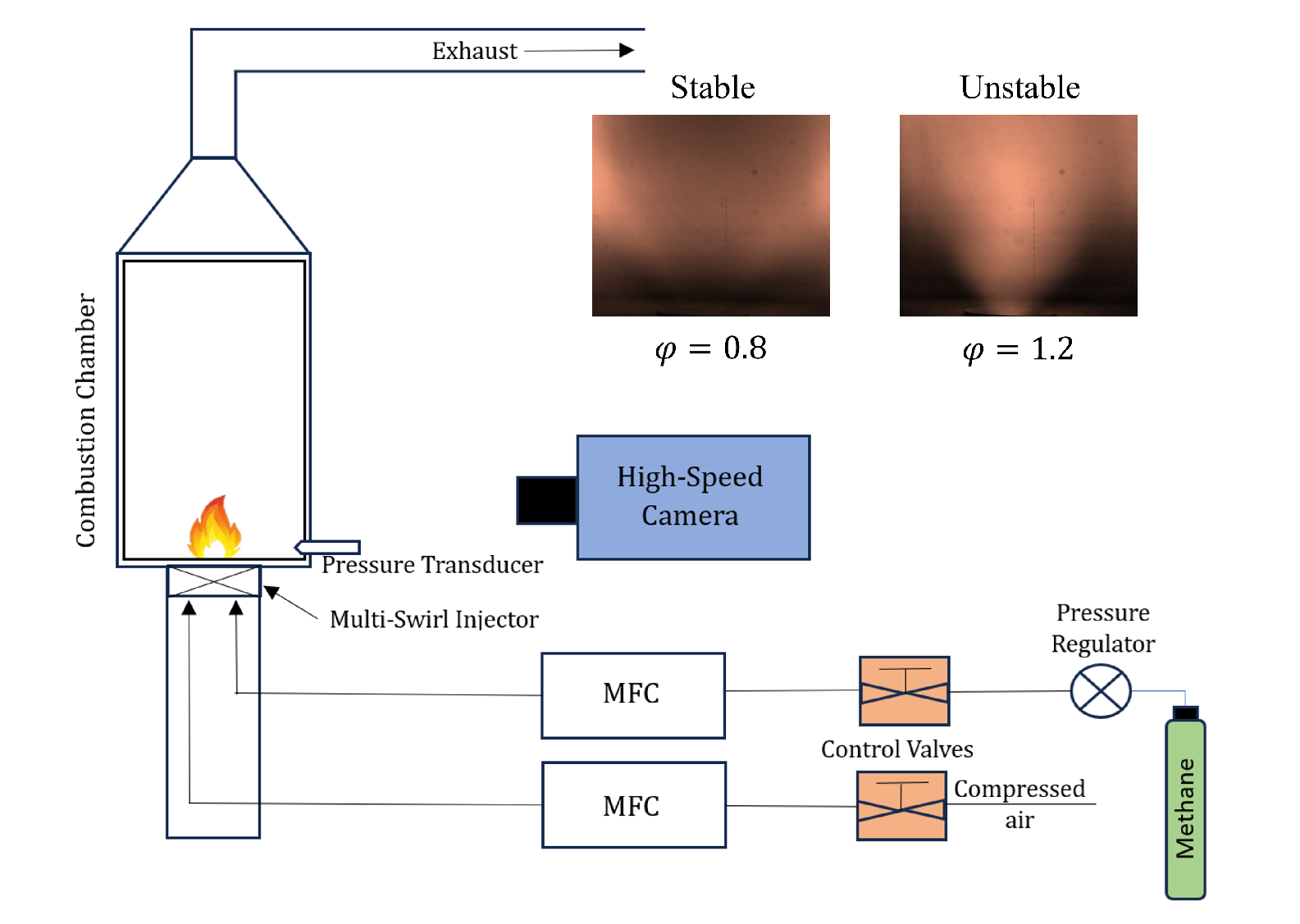}
    \caption{Schematic of the experimental setup with flame stabilized at the dump plane using counter-rotating high shear swirl injector. The air and fuel flow lines are shown for clarity along with flame images (insets).} 
    \label{fig_exp_setup}
\end{figure}

\section{\uppercase{Experimental setup, data acquisition and combustor dynamics}}\label{sec2_exp}
Combustion experiments were carried out on a laboratory scale model can gas turbine combustor. Figure \ref{fig_exp_setup} shows the schematic of combustor with flow lines and instruments marked. Counter-rotating high-shear swirl injectors are used for fuel air mixing and subsequent flame stabilization. The counter swirl configuration supplies the air with 60:40 in the clock and counter clock wise directions, respectively, with geometric swirl number of the primary and secondary swirlers are 1.5 and 0.8, respectively. Methane (CH$_4$) is used as fuel and is supplied from the high pressure cylinder using pressure regulator. The fuel flow rate is controlled using Alicat mass flow controller. Here, only the fuel flow is altered to change the equivalence ratio $[0.80-1.20]$ with a step size of 0.10. The air is supplied at a constant airflow rate (300 slpm) with a Reynolds number of 14708 from a high pressure tank. The air flow is also controlled using a Alicat mass flow controller. The air and fuel supply lines are shown in Figure \ref{fig_exp_setup}. Since, the fuel and air meets inside the swirl chamber, the combustor operates in a technically premixed mode.

\begin{figure} [h]
\begin{minipage}[b]{0.6\textwidth}
\hspace{5mm}
\begin{subfigure}{0.9\textwidth}
\hspace{-10mm}
\includegraphics[width=0.48\linewidth]{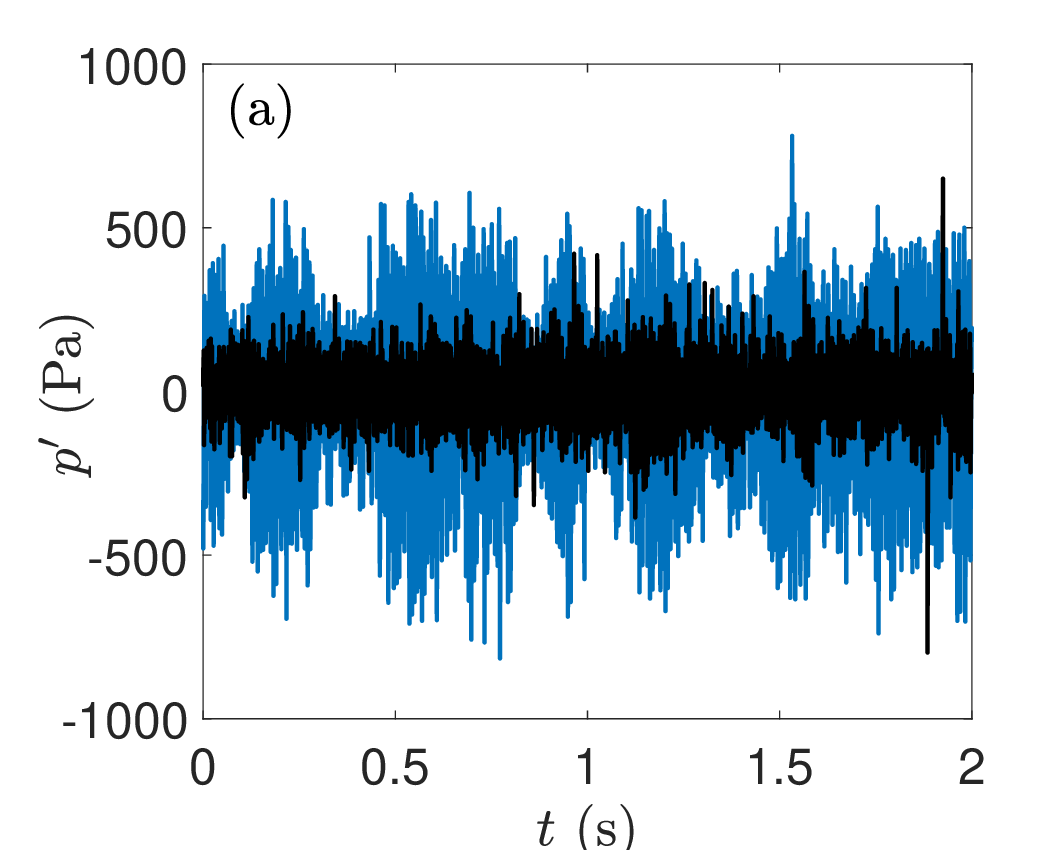}
\hspace{-4mm}
\includegraphics[width=0.48\linewidth]{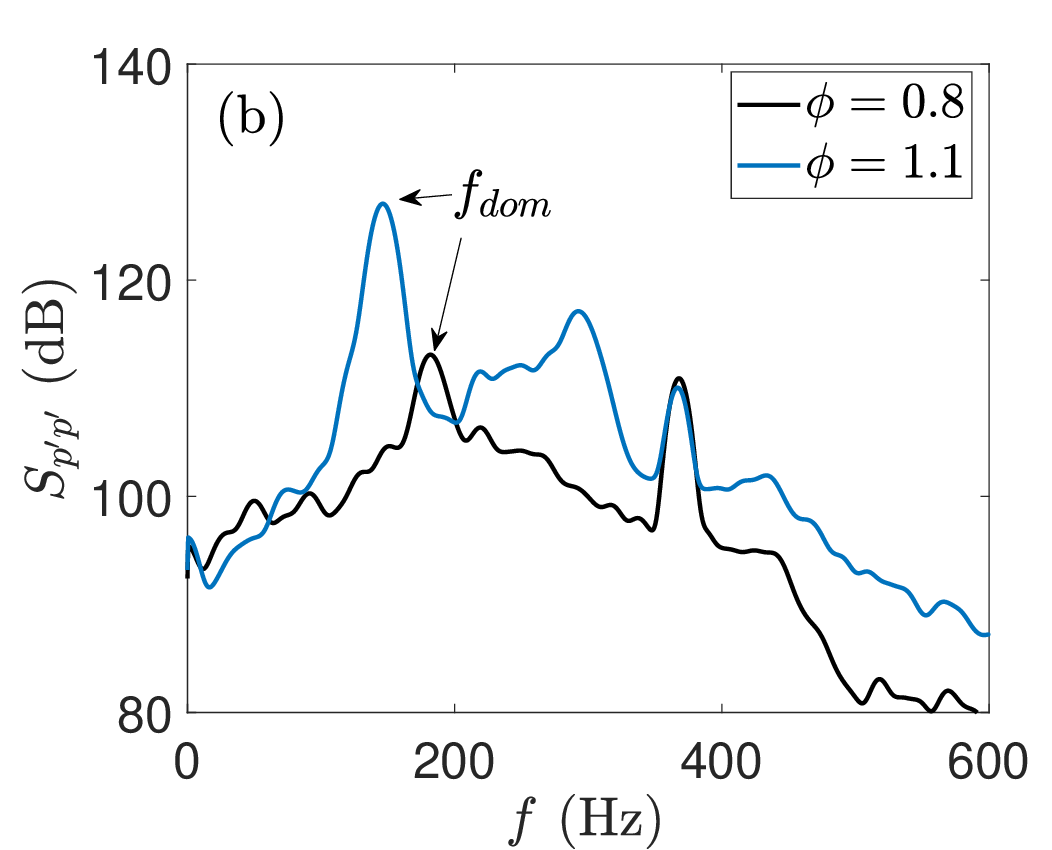}
\end{subfigure}
\end{minipage}
\quad
\caption{(a) Acoustic pressure and associated (b) sound pressure level for stable (black solid line) and unstable (blue solid line) operations, respectively. $f_{dom}$ denotes the dominant peak associated with duct acoustic modes.}
	\label{fig_pr_ts_SPL}
\end{figure}

The acoustic pressure ($p^\prime$) in the combustion chamber is recorded using a dynamic transducer located near the dump plane as shown in Fig. \ref{fig_exp_setup}. The $p^\prime$ is sampled at a frequency of 20 kHz and NI 9205 module installed in an NI cDAQ 9179 chassis for data acquisition. In addition, high-speed flame images are captured using Photron SA5 high speed CMOS camera.

The acquired acoustic pressure and associated power spectral densities are shown in Fig. \ref{fig_pr_ts_SPL}(a) and (b), respectively, to clarify the instability dynamics. The black and blue solid lines corresponds to equivalence ratios of $0.8$ and $1.1$ labeled as stable and unstable operation. The unstable operation exhibit relatively large sound pressure level (127 dB) at discrete frequency of 146 Hz. On the other hand, the stable operation, the sound pressure level lies in the order of back ground noise.

\begin{figure}
	\centering
	\includegraphics[scale=0.355]{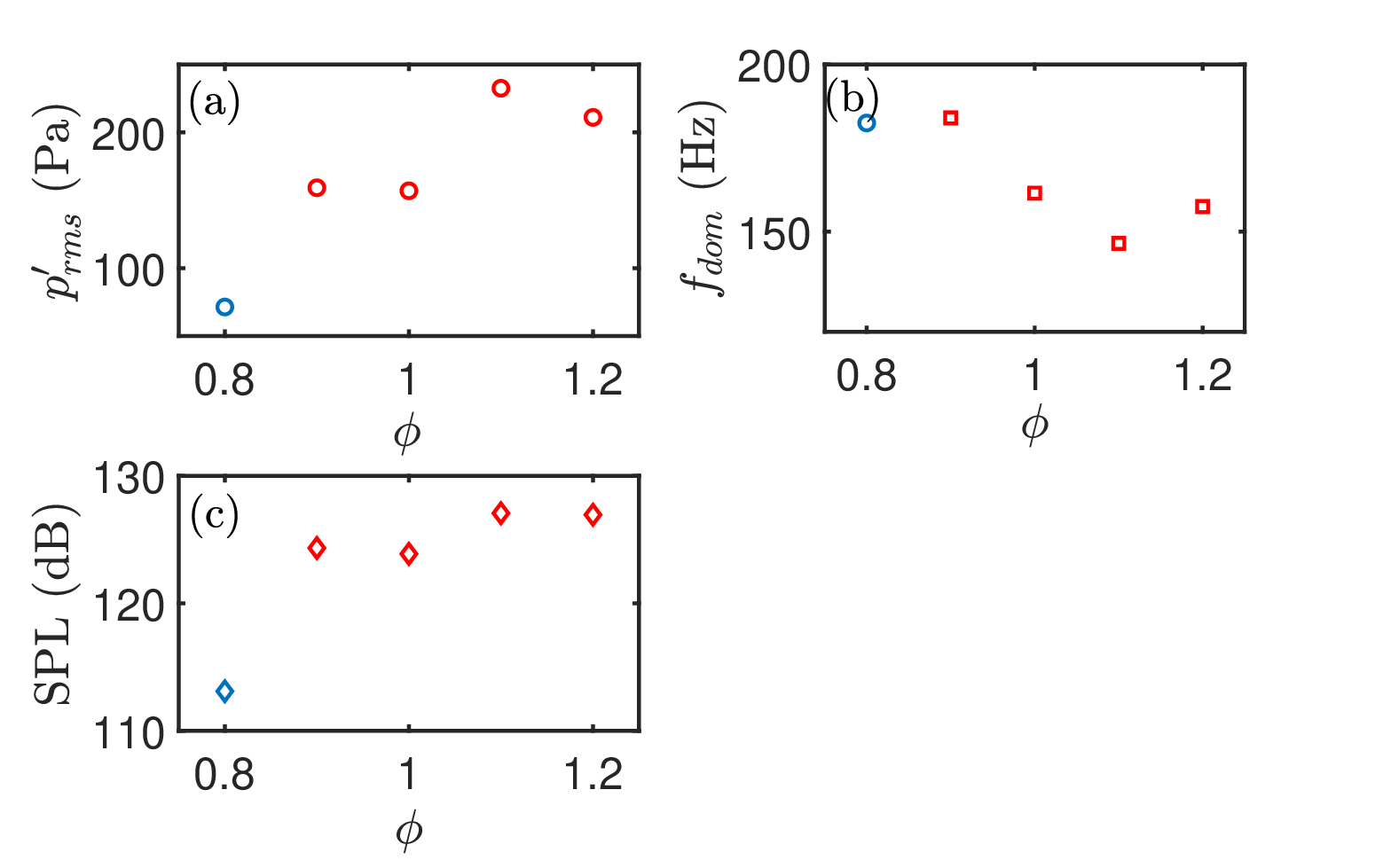}
	\caption{(a) Root mean square value of the acoustic pressure, (b) dominant instability frequency and associated (c) sound pressure level (SPL) across various $\phi$. The blue and red markers denote stable and unstable operating conditions.}
	\label{fig_prms_fdom_SPL}
\end{figure}

Figure \ref{fig_prms_fdom_SPL}(a) shows root mean square value of acoustic pressure for various equivalence ratios explored. Apart from $\phi=0.80$, the other operating conditions exhibit relatively large amplitude acoustic pressure oscillations. In addition, the $f_{dom}$ marked in Fig. \ref{fig_pr_ts_SPL}(b) is plotted for other $\phi$ for clarity in Fig. \ref{fig_prms_fdom_SPL}(b). It is observed that for large amplitude acoustic oscillations ($p^\prime_{rms}$) lead to relatively reduction in instability frequency ($f_{dom}$). Further, in Fig. \ref{fig_prms_fdom_SPL}(c) the sound pressure level (SPL) associated with $f_{dom}$ are shown and the variation with $\phi$ is similar to $p^\prime_{rms}$. The stable and unstable operating conditions in Fig. \ref{fig_prms_fdom_SPL}(a-c) are marked with blue and red markers, respectively, for clarity. The following section briefly discusses about the measures considered in the present study for combustion condition classification. In the present study for combustion condition identification, we classify the $\phi=0.80$ as a stable condition and other operating conditions as unstable.

\section{\uppercase{Methodology}}\label{methodology}
In this section, we briefly discuss about the measures used in suitably identifying the combustion condition.

\subsection{Root means square (RMS) values}
The root mean square value of the acoustic pressure is the root of arithmetic mean of the square of the individual values in a given time window. The mathematical expression is as follows.
\begin{equation}
    p^\prime_{rms} = \left[\frac{1}{N}\sum_{i=1}^{N} p^{\prime_i 2} \right]^{1/2}
\end{equation}
where, $p^\prime_i$ is the acoustic pressure sample at $i$ and $N$ is the total data length in a window considered. 

\subsection{Signal to noise ratio (SNR) }
Signal to noise ratio is the ratio of the strength of the desired signal to the back ground noise. Therefore, the high and low values of SNR indicates combustion instability and stable combustion, respectively. In the present study, the thermoacoustic instability occurs in the frequency range of $145\le f_{dom}\le185$ Hz. The average background noise are measured within the frequency range of $500 \le f \le 600$ Hz, as there are no higher harmonics present in this range. The SNR used in the present study can be expressed as follows.
\begin{equation}
    SNR =   \frac{A(f_{dom})}{ \bar{A}(500\le f_i\le600) }
\end{equation}

\subsection{Phase space Reconstruction}
Phase space reconstruction proves to be an important tool for the visualization of the dynamics of a system. It is difficult to extract conclusions from the time series data plotted on one dimension. The data is mapped to a high-dimensional space where the trajectories evolve with time, and the study of these trajectories provides meaningful insights into combustion dynamics. It also provides a good insight into the chaotic, periodic and aperiodic nature of the attractor.

The construction of phase space involves extraction of time delay ($\tau$) and embedding dimension ($d$) of the dynamical system which are computed using Average Mutual Independence (AMI) and False Nearest Neighbours (FNN) algorithms, respectively. 
\subsubsection{Average Mutual Information (AMI)}
Mutual information of two variables quantifies the dependence of two variables on each other. This method is used to find the optimal time delay for the time series data. While constructing the phase space, it is important for the variables to be independent. Hence, fairly large time delays must be chosen. However, large time delays can lead to highly uncorrelated variables which cannot represent the dynamics of the system. For each time delay, the average mutual information is computed, and the first minima of the AMI vs time delay is taken to be the optimal time delay for phase space reconstruction. The formula for average  mutual information is as follows. 
 \begin{equation}
    I(\tau)=\sum_{p(t_{i}),p(t_{i}+\tau)}{P(p(t_i),p(t_{i}+\tau))}\log_{2}{\left[\frac{P(p(t_{i}),p(t_{i}+\tau))}{P(p(t_{i}))P(p(t_{i}+\tau))}\right]}
 \end{equation}
  
\subsubsection{False Nearest Neighbors (FNN)}
The embedded dimension is computed using the false nearest neighbours algorithm. If a system lies in n-dimensional space in reality but we assume it to be lower, then the attractor trajectory winds and crosses itself, which might not have happened in the higher-dimensional space. This can potentially lead to false notions of the system. 

If there is an increase in the Euclidean distance between neighbourhood points  as the number of the embedded dimensions increases by one, then the neighbourhood points are not the true neighbours of the point. This implies that the system lies in a higher dimension than assumed. The embedding dimensions are increased, and the percentage of false neighbours is calculated for each embedding dimension. The dimension for which the percentage of false neighbours falls below a certain value for the first time is taken as the embedding dimension of the system.

The distance between the point of interest and its $r^{th}$ neighbour in $d-$dimensions is given as
\begin{equation}
    R_d^2(n,r)=\sum_{k=1}^{d}[x(n+kT)-x^{(r)}(n+kT)]^2
\end{equation}
The corresponding distance in $d+1$ dimension is given as
\begin{align}
    R_{d+1}^{2}(n,r) & =\sum_{k=1}^{d+1}{[x(n+kT)-x^{(r)}(n+kT)]^2}\\
    & =R_{d}^{2}(n,r)+[x(n+kT)-x^{(r)}(n+kT)]^2.
\end{align}

The criterion for determining false neighbourhood point is
\begin{equation}
    \frac{[R_{d+1}^{2}(n,r)-R_{d}^{2}(n,r)]^{\frac{1}{2}}}{R_{d}(n,r)} > R_{T}.
\end{equation}
Where, $R_{T}$ is the threshold distance to regard a point as a false neighbour.

\subsection{Fractal Dimension (FD)} 

Box-counting is the simple yet efficient method used in determining the fractal dimension based on the number of boxes whose dimension tends to zero used to cover the curve. This is formalized as
\begin{equation}
    D=\lim_{\epsilon \to 0}\left[\frac{\log(N(\epsilon))}{\log(1/\epsilon)}\right]
\end{equation}
where, $N(\epsilon)$ is the number of boxes of side $\epsilon$ needed to cover the object.

\subsection{Hurst Exponent}
The Hurst exponent is the measure of the long-term memory of the time series. It can also be seen as the degree of self-similarity in the time series. It is defined as 
\begin{equation}
    \mathbb{E}\left[\frac{R(n}{S(n)}\right]=Cn^{H}
\end{equation}
where, $R(n)$ is the range of the cumulative deviations for $n$ segments of the data and $S(n)$ is the standard deviations of the data for each segment. The ratio $\frac{R(n)}{S(n)}$ is the re-scaled range averaged over all the data segments. $\mathbb{E}$ denotes the expected value, $C$ is a constant, and $H$ is the Hurst exponent, which is calculated by fitting a linear model to the log-log plot of the re-scaled range and the data segments.

\subsection{Recurrence Plot}
A recurrence plot is another important tool in time-series analysis, which facilitates visualisations of the complex attractor trajectories that exist in higher dimensional space. It is a time versus time plot and, hence, is a symmetric plot. It is formalized as
\begin{equation}
    R_{i,j} = \Theta(\epsilon - ||\vec{x_{i}} - \vec{x_{j}}||)
\end{equation}
 where, $\Theta$ is the Heaviside function.

If the attractor trajectory revisits some point in phase space at times $t_1$ and $t_2$ within a distance threshold of $\epsilon$, then the $t_1$, $t_2$ and the $t_2$, $t_1$ entries of the recurrence matrix would be 1. This produces a 2-dimensional matrix of zeroes and ones, which is then visualized by a binary image. These plots give insights into the dynamics of the system through its geometrical characterization.

Based on the above insights following features can be defined.
\begin{enumerate}

    \item \textit{Trapping time (TT)}: This quantifies the average of vertical/horizontal lengths that are more than a specified threshold ($H_{min}$) in the recurrence plot.
    \begin{equation}
        TT = \frac{\sum_{H_{min}}^{N}{hP(h)}}{\sum_{H_{min}}^{N}{P(h)}}
    \end{equation}
    \item \textit{Laminarity (LAM)}: It quantifies the fraction of horizontal/vertical lines that are higher than a specific threshold ($H_{min}$).
    \begin{equation}
        LAM = \frac{\sum_{H_{min}}^{N}{hP(h)}}{\sum_{1}^{N}{hP(h)}}
    \end{equation}
\end{enumerate}

\begin{figure}
\begin{minipage}[b]{0.6\textwidth}
\hspace{5mm}
\begin{subfigure}{0.9\textwidth}
\subcaption{}
\hspace{-11mm}
\includegraphics[width=0.5\linewidth]{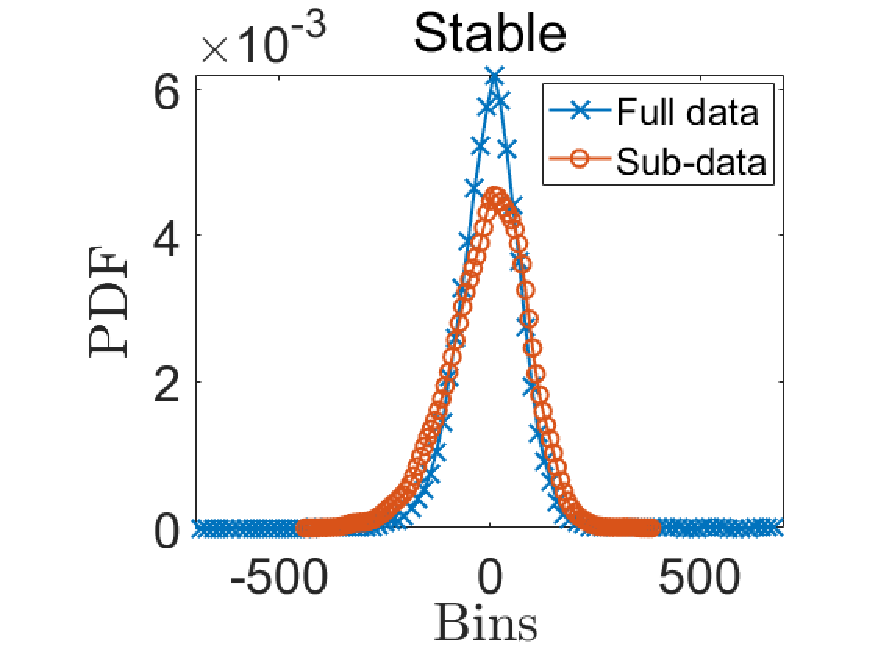}
\hspace{-6mm}
\includegraphics[width=0.5\linewidth]{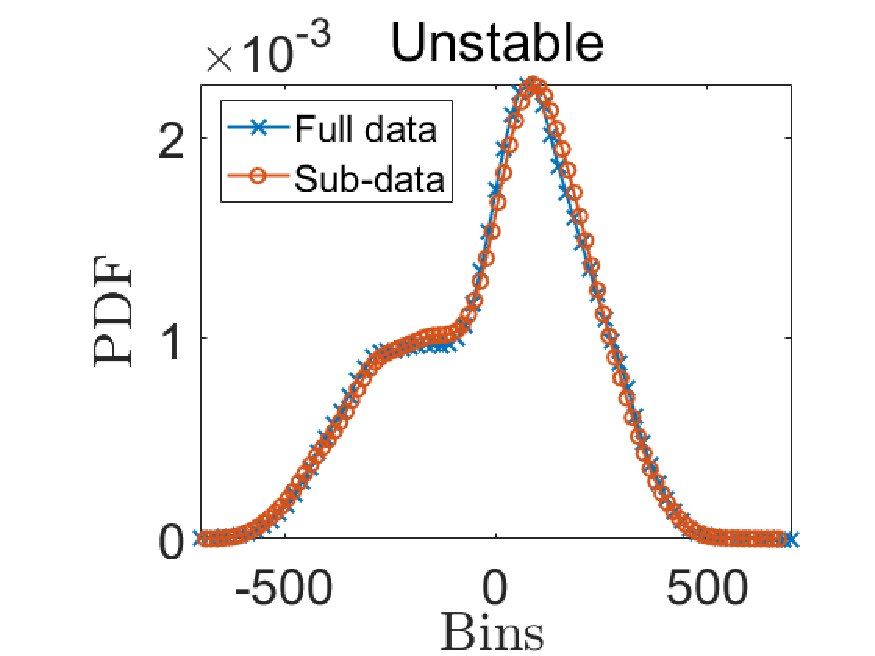}
\end{subfigure}
\end{minipage}

\begin{minipage}[b]{.6\textwidth}
\hspace{5mm}
\begin{subfigure}{0.9\textwidth}
\subcaption{}
\hspace{-8mm}
\includegraphics[width=0.48\linewidth,height=3.875cm]{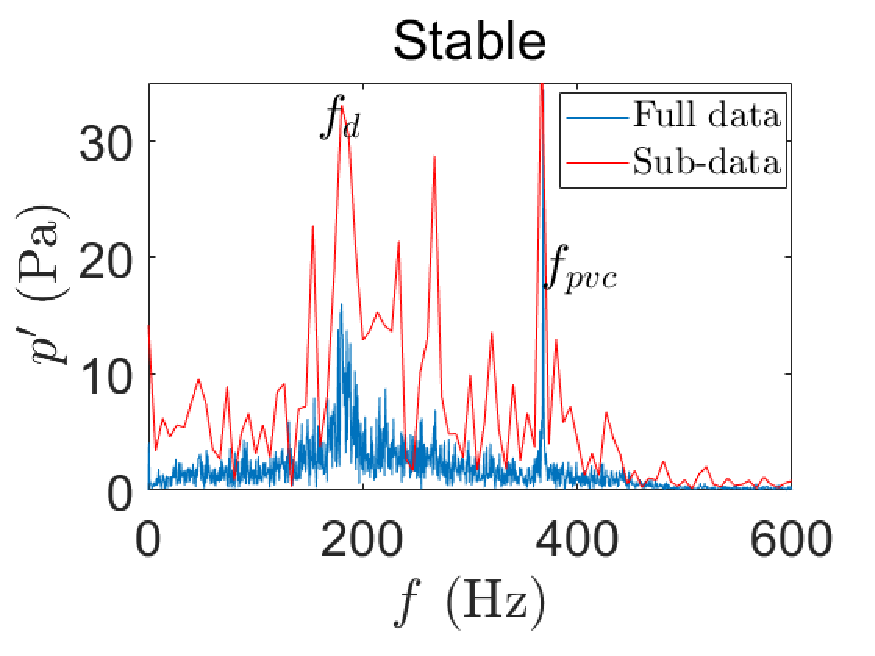}
\hspace{-4mm}
\includegraphics[width=0.49\linewidth]{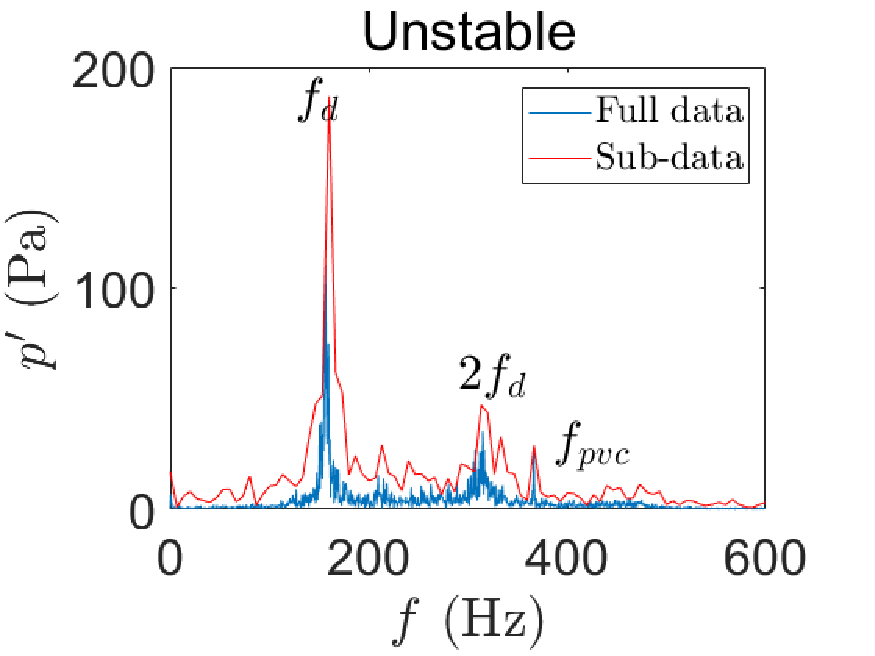}
\end{subfigure}
\end{minipage}

\quad
\caption{Comparison of (a) PDF and (b) Fourier spectrum between full data (blue) and sub-data (red). Left and right columns correspond to stable and unstable conditions.}
\label{fig_FFT_PDF}
\end{figure}

\section{\uppercase{Machine Learning (ML) Model}}
\label{ML_model}
The machine learning is broadly divided into supervised, unsupervised and reinforcement learning. Supervised learning is based on classification of the labelled data where we know which class each of the data belongs to. The number of classes are well defined. Whereas unsupervised learning deals with classification of the data whose classes are not known. In this paradigm, the model tries to discover patterns in the data and classify based on these patterns without human supervision. Reinforcement learning is based on trial-and-error type of learning where the model tries to learn as it makes predictions. The present study uses supervised learning.

Binary classification is part of the supervised learning where the model is trained to classify the data into two classes (for instance, stable and unstable). Various ML models exist for binary classification like Decision Trees (DT), Random Forest (RF), Support Vector Machines (SVM), and Artificial Neural Networks (ANN), etc. We resort to using DT owing to its simplicity, low-computational cost, and reasonably good predictions. 

\subsection{Data processing and feature extraction}
\begin{figure}
    \begin{subfigure}[t] {0.5\textwidth}
        \includegraphics[width=0.47\linewidth]{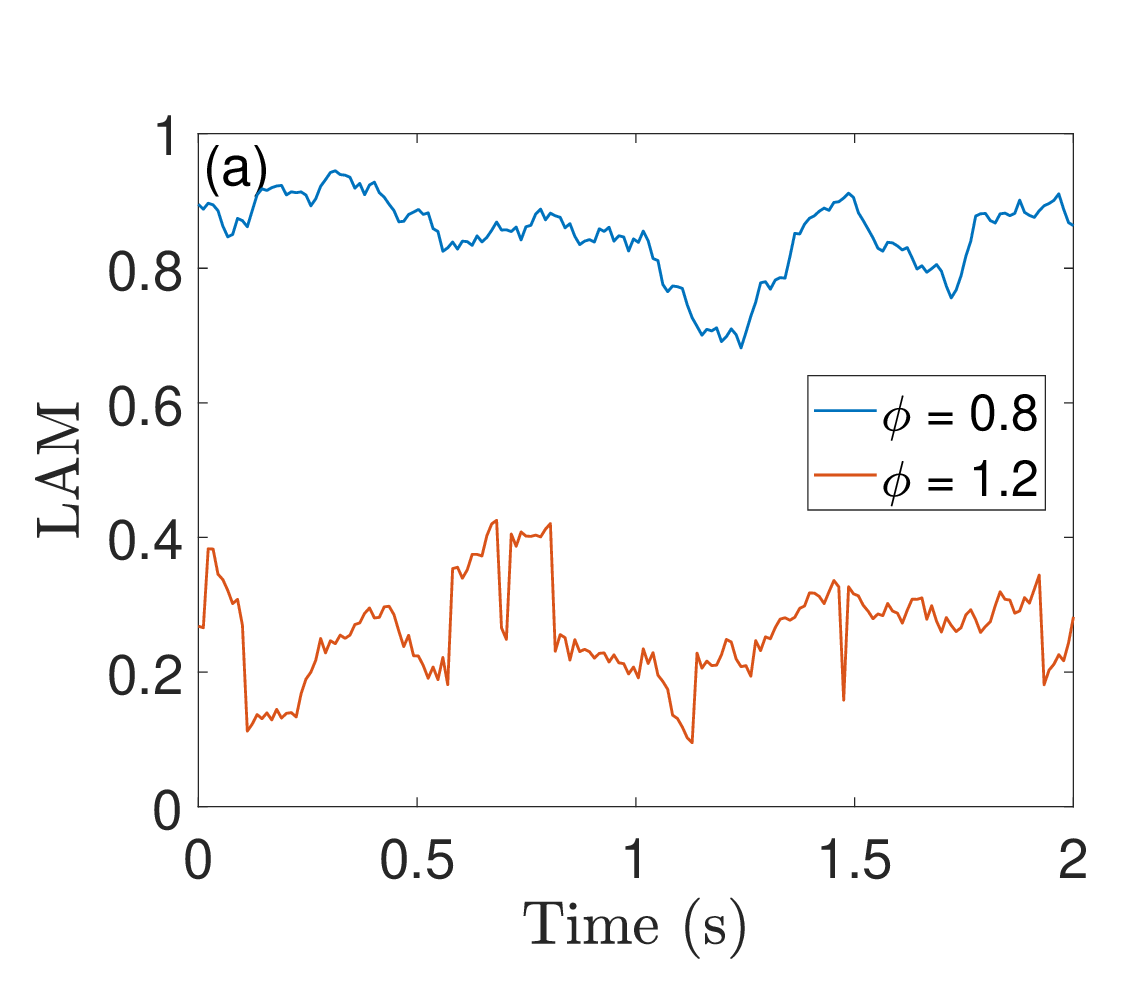}
        \hspace{0.2cm}
        \includegraphics[width=0.47\linewidth]{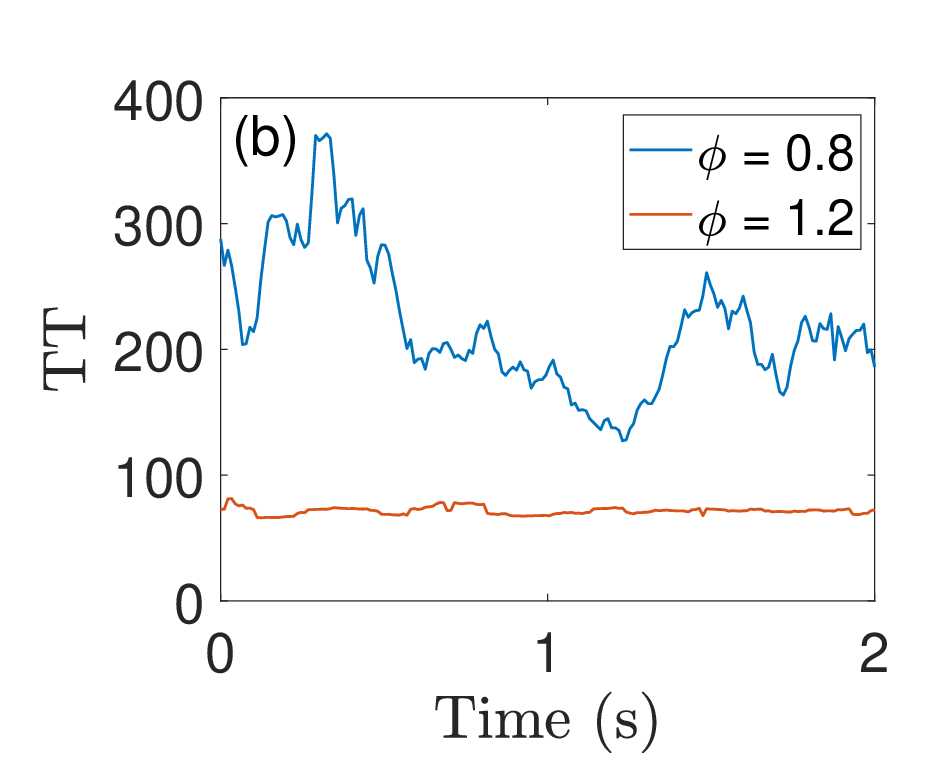}
        
    \end{subfigure}
    \vspace{0.5cm}
    \begin{subfigure}[t] {0.5\textwidth}
        \includegraphics[width=0.5\linewidth]{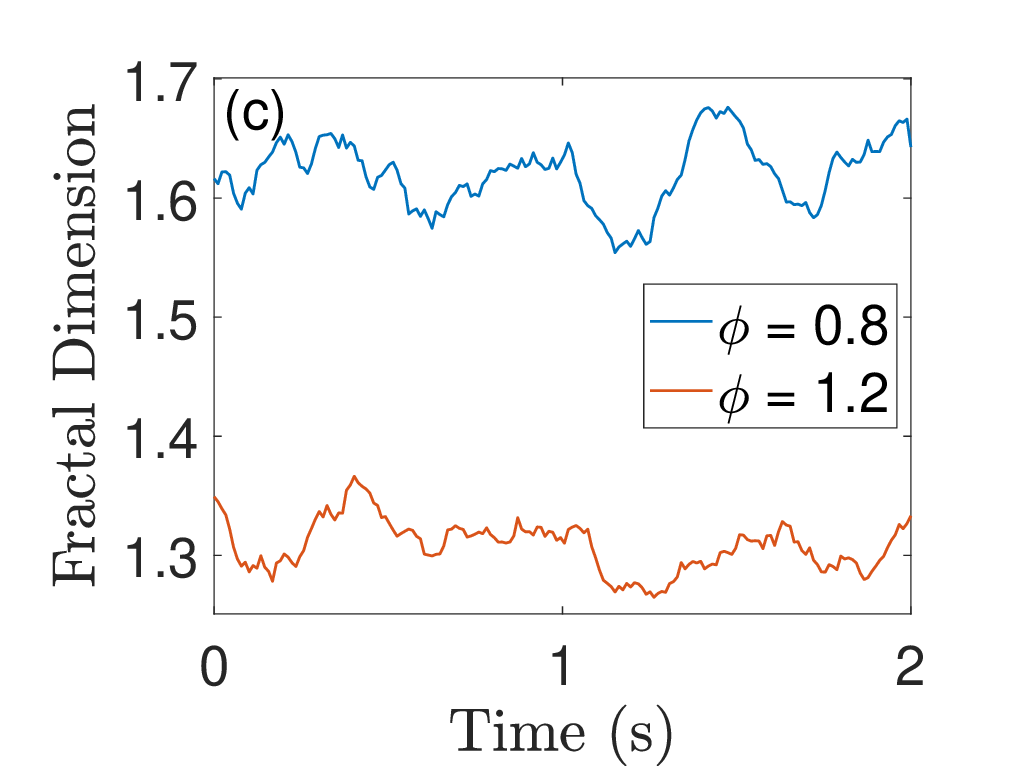}
        \includegraphics[width=0.5\linewidth]{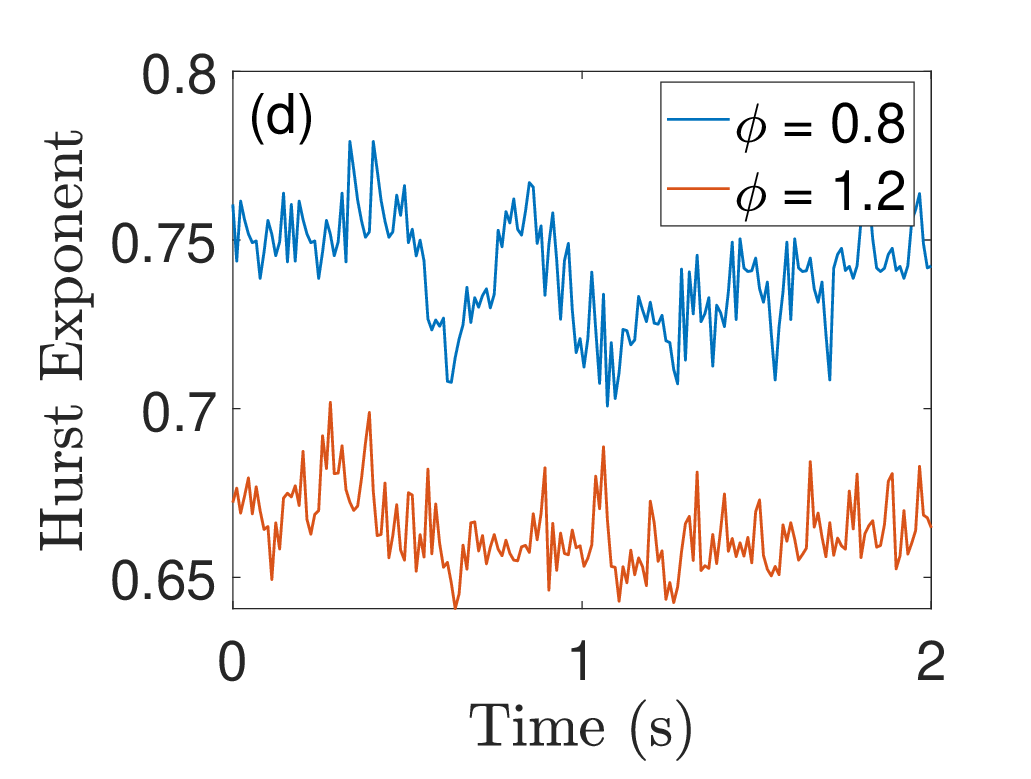}
        
    \end{subfigure}
    \caption{(Variation of (a) $LAM$, (b) $TT$, (c) fractal dimension, and (d) Hurst exponent with time, respectively.}
    \label{fig_Measure_Evolu}
\end{figure}

As mentioned in section \ref{sec2_exp}, we have the  acoustic pressure time series for various equivalence ratios, each time series containing 40,000 data points (2 seconds). Each time series data is divided into several sub-data. The sub-data length is evaluated by comparing the fast Fourier transform and probability distribution of sub-data with that of the over all data. This comparisons are shown in figure \ref{fig_FFT_PDF}. We found good match in spectral contents and acoustic pressure distributions for 3000 data points. Therefore, a data length of 3000 is chosen for sub data length. This data length contains close to 30 dominant acoustic pressure cycles. This justifies the choice of sub-data length in this study. A sliding window is implemented with a data length of 3000 and increments of 150 is used to extract features discussed in section \ref{methodology}. These are shown in figure \ref{fig_Measure_Evolu} for stable ($\phi=0.8$) and unstable ($\phi=1.2$) combustion operating conditions for clarity.

In fig. \ref{fig_Measure_Evolu}, the blue and red lines correspond to stable and unstable combustion conditions. The panels (a,b) show the time evolution of laminarity (LAM) and trapping time (TT) calculated from recurrence plot for whole data points. The relatively coherent fluctuation behaviour of unstable condition features relatively low value of LAM and TT compared with random nature of turbulence and combustion noise associated with the stable condition. Similarly, the fractal dimension and Hurst exponents are calculated and shown in panels (c,d), respectively. Again, the stable conditions features relatively large value of FD and HE than unstable condition. Overall, from panels (a-d), it is observed that the stable (blue line) and unstable (red line) combustion conditions can be clearly separeted by the parameters considered for classification. A similar measure, $p^\prime_{rms}$ and signal-to-noise (SNR) ratio is calculated for subsequent classification for all the operating conditions discussed in section \ref{sec2_exp}.

\subsection{ML model architecture and training}
Decision trees are used for the binary classification of stable and unstable combustion condition in this study. The architecture involves partitioning the feature space so as to differentiate one class of data from the other. For ease, we assign a numeric values for the labels stable and unstable to 1 and 0, respectively. The partitioning of the feature space is done by asking a "question" at each decision node used for partitioning of the dataset in the next level of nodes. Once the dataset is partitioned, impurity of the partitioned dataset is analysed which tells the quality of our partitioning. This impurity is quantified using entropy ($E$): 
\begin{equation}
    E(P) = -(P\log{P} + (1-P)\log{(1-P)}).
\end{equation}
Here, $P$ is the fraction of correctly classified data at a particular node. The goodness of the question is then quantified using the entropies calculated at a node and its child nodes. The quantity is called the Information Gain ($IG$) and described as follows.

\begin{equation}
    IG = E(P) - [\gamma E(P_1) + (1-\gamma)E(P_2)]
\end{equation}
where, $P_1$ and $P_2$ represents the fraction of correctly classified data at two child nodes. $\gamma$ is the ratio of fraction of correctly classified data points in the first child node to the fraction of correctly classified data points in the parent node. Hence, the "question" is decided in each node such that IG is maximum.

It is important to note that as, we grow the tree long, there is a high possibility of over-fitting the data. This might lead to bad performance on the testing data such as training the outliers also. Hence, tree growth should be terminated based on purity of classification on the leaf nodes. 

For simplicity, we used three sets of two features each in training the model such as RMS of acoustic pressure and signal-to-noise ratio, Hurst exponent and fractal dimension, laminarity and trapping time. In the present study, we found these features to be most suitable for classification compared to other features discussed (not shown here) in section \ref{sec2_exp}. Three decision trees were trained with each of the above set of features. The model was trained with $\phi = 0.8$ and $\phi = 1.2$ operating conditions. The data for other operating conditions ($\phi = 0.9,~ 1 ~\& ~1.1$) was used for testing the model. The classifier testing, validation and its accuracy are discussed in the next subsection.

\subsection{Classifier testing, validation and accuracy}
Data for two operating conditions ($\phi = 0.8$ and 1.2) are used for training the model. The data used for training is subdivided using a K-fold validation scheme. In this scheme, the data is divided into two parts of sizes, $N/k$ and $N(1-1/k)$, where N is the total number of data points and k is an integer. The decision tree algorithm is run on the bigger subset of the data $N(1-1/k)$ and tested on the smaller subset $N/k$. This process of training is done for different subsets of data. Here, 5-fold validation scheme is used for training the decision tree where model undergoes training 5 times with $80\%$ training data and $20\%$ testing data at each training step.

\begin{figure}
    \centering
        \centering
        \includegraphics[width=0.49\linewidth]{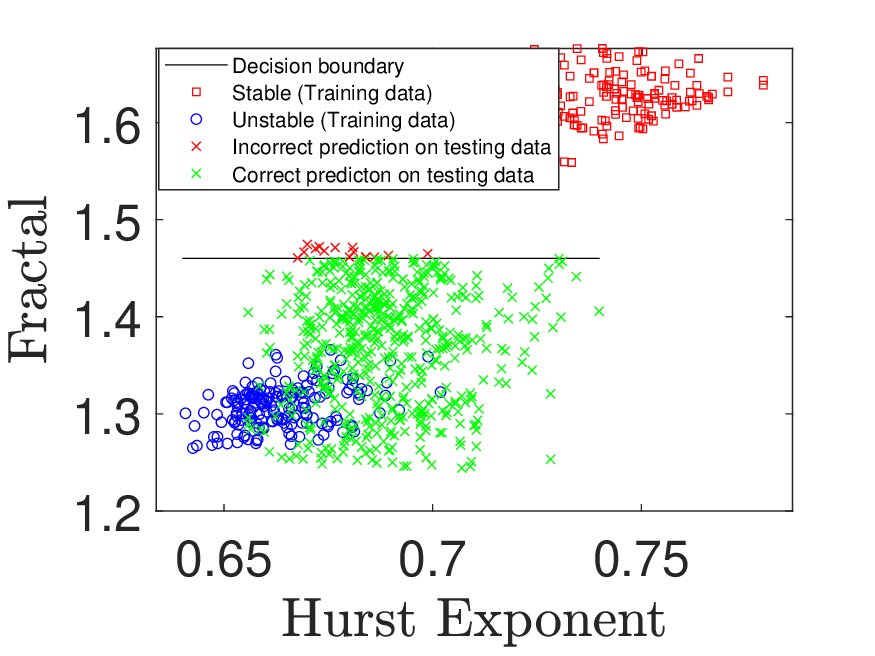}
        \centering
        \includegraphics[width=0.49\linewidth]{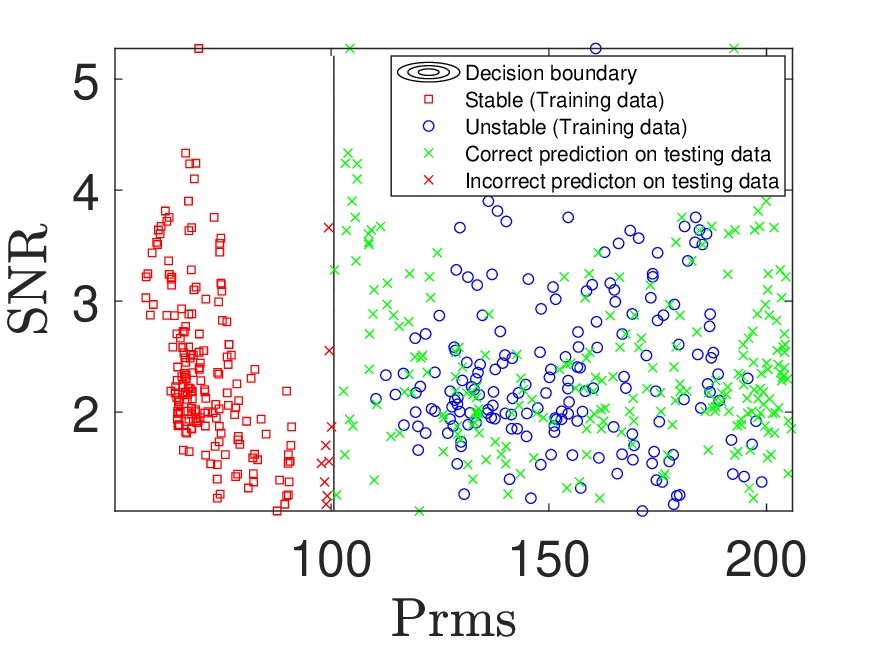}
        \centering
        \includegraphics[width=0.5\linewidth]{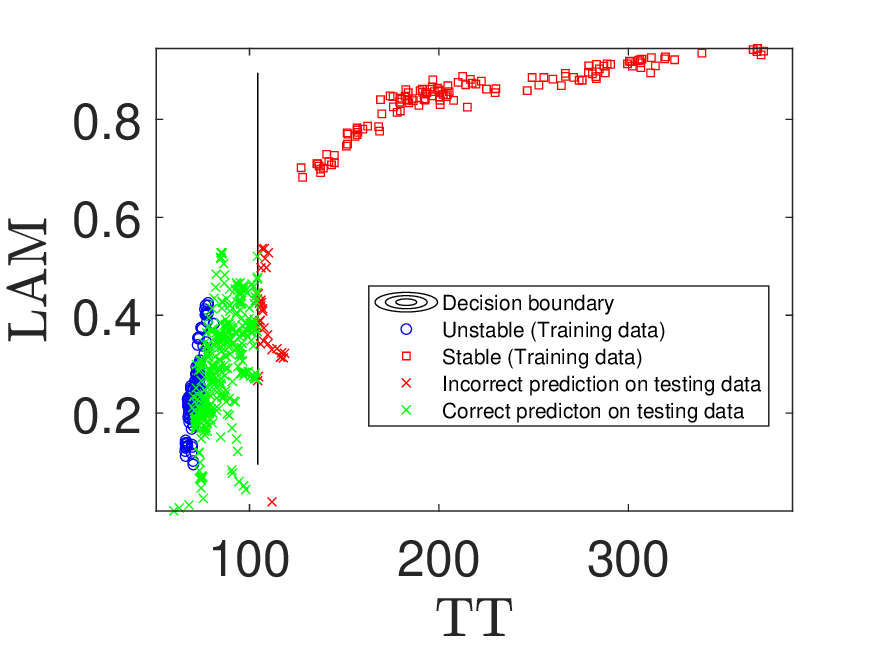}
    \caption{Scatter plot of the stable and unstable data points with decision boundaries for three decision tree model with different set of features:     (a) HE vs FD (b) $P_{rms}$ vs $SNR$, (c) LAM vs TT.}
    \label{fig_scatter_plot}
\end{figure}

The model trained using K-fold validation is then tested using the data of the other operating conditions. The scatter plots and the decision boundaries can be see in Fig. \ref{fig_scatter_plot}. In Fig. \ref{fig_scatter_plot}, the blue and red symbols denote stable and unstable operating conditions, whereas the decision boundary is marked by black lines. Clustering of the points and  their clear separation of the stable and unstable combustion condition is discernible. Table \ref{tab_model_accuracy} shows the performance of different decision tree models associated with the various features considered on training and testing the data.

\begin{table}
    \centering
    \begin{tabular}{|c|c|c|}
    \hline
    Models & Training set & Test set \\
    \hline
    Model 1 & 99.2 & 97.22 \\
    Model 2 & 100 & 98.33 \\
    Model 3 & 100 & 90.86 \\
    \hline
    \end{tabular}
    \caption{Decision tree model accuracy. Model 1: $P_{rms}$ vs $SNR$, Model 2: HE vs FD, and Model 3: LAM and TT.}
    \label{tab_model_accuracy}
\end{table}

\section{\uppercase{Conclusions and future direction}}\label{conclusion}
In this study, we applied a machine learning-based binary classification model to distinguish between stable and unstable operations of a combustor using a high shear swirl injector, commonly found in modern gas turbine engines due to its superior atomization properties. Identifying dynamic signatures of such combustors is critical for mitigating combustion instability. Acoustic pressure and high-speed flame imaging were collected in a model swirl combustor at atmospheric conditions, with the equivalence ratio varied at a fixed Reynolds number. The dominant thermoacoustic instability frequency observed in the range 145-185 Hz. Features like root mean square value, signal-to-noise ratio, laminarity, trapping time, Hurst exponent, and fractal dimension were extracted using time series analysis. These features formed clusters corresponding to stable and unstable operations. A decision tree-based machine learning model was then trained and tested, demonstrating its effectiveness in identifying combustion conditions in the high shear injector combustor which is of practical relevance.

\section*{\uppercase{Acknowledgements}}
This work was carried out by PKA (undergraduate student) during his two month  summer research internship program at Indian Institute of Science Bangalore.

\section*{\uppercase{references}}
\begingroup
\renewcommand
\refname{}
\bibliographystyle{elsarticle-num}
\bibliography{bibliography}
\endgroup

\end{document}